\documentclass[conference]{IEEEtran}
\IEEEoverridecommandlockouts
\usepackage{cite}
\usepackage{amsmath,amssymb,amsfonts}
\usepackage{graphicx}
\usepackage{textcomp}
\usepackage{xcolor}
\def\BibTeX{{\rm B\kern-.05em{\sc i\kern-.025em b}\kern-.08em
    T\kern-.1667em\lower.7ex\hbox{E}\kern-.125emX}}

\usepackage{algorithm}
\usepackage{algorithmicx}
\usepackage[noend]{algpseudocode}

\usepackage{multirow}
\usepackage{hyperref}
\usepackage{subcaption}

\begin{document}

\title{Urban Water Consumption Forecasting Using Deep Learning and Correlated District Metered Areas
}



\author{
	\IEEEauthorblockN{
		Kleanthis Malialis\textsuperscript{a},
		Nefeli Mavri\textsuperscript{a},
		Stelios G. Vrachimis\textsuperscript{a},
		Marios S. Kyriakou\textsuperscript{a},\\
		Demetrios G. Eliades\textsuperscript{a} and
		Marios M. Polycarpou\textsuperscript{a, b}
	}
	\IEEEauthorblockA{
		\textsuperscript{a} \textit{KIOS Research and Innovation Center of Excellence}\\
		\textsuperscript{b} \textit{Department of Electrical and Computer Engineering}\\
		\textit{University of Cyprus},
		Nicosia, Cyprus\\
            Email: \{malialis.kleanthis, mavri.nefeli, vrachimis.stelios, kiriakou.marios, eldemet, mpolycar\}@ucy.ac.cy
        \thanks{ORCID: 0000-0003-3432-7434 (KM), 0000-0001-8862-5205 (SGV), 0000-0002-2324-8661 (MSK), 0000-0001-6184-6366 (DGE), 0000-0001-6495-9171 (MMP)}
        \thanks{This paper was supported by the European Research Council (ERC)
under grant agreement No 951424 (Water-Futures), the European Union’s
Horizon 2020 research and innovation programme under grant agreement No
739551 (KIOS CoE), the Republic of Cyprus through the Deputy Ministry of
Research, Innovation and Digital Policy.}
        }
}

\maketitle

\begin{abstract}
Accurate water consumption forecasting is a crucial tool for water utilities and policymakers, as it helps ensure a reliable supply, optimize operations, and support infrastructure planning. Urban Water Distribution Networks (WDNs) are divided into District Metered Areas (DMAs), where water flow is monitored to efficiently manage resources. This work focuses on short-term forecasting of DMA consumption using deep learning and aims to address two key challenging issues. First, forecasting based solely on a DMA’s historical data may lack broader context and provide limited insights. Second, DMAs may experience sensor malfunctions providing incorrect data, or some DMAs may not be monitored at all due to computational costs, complicating accurate forecasting.
We propose a novel method that first identifies DMAs with correlated consumption patterns and then uses these patterns, along with the DMA's local data, as input to a deep learning model for forecasting. In a real-world study with data from five DMAs, we show that: i) the deep learning model outperforms a classical statistical model; ii) accurate forecasting can be carried out using only correlated DMAs' consumption patterns; and iii) even when a DMA's local data is available, including correlated DMAs' data improves accuracy.
\end{abstract}


\section{Introduction}\label{sec:intro}

\subsection{Motivation}
A District Metered Area (DMA) is a well-defined and isolated zone within an urban Water Distribution Network (WDN), where water in-flow is continuously monitored. Accurate forecasting of water demand in each DMA allows city planners and utility managers to ensure a reliable supply, mitigating the risk of shortages during peak demand periods. 
DMA flow/consumption forecasting is also integral to the development of Digital Twins \cite{Fuertes2020}, that enable water utilities to optimize their operations, manage resources more efficiently, and detect operational abnormalities in a timely manner. Moreover, water demand forecasting plays a key role in strategic infrastructure planning, helping guide the expansion of water treatment plants and the upgrade of distribution networks to accommodate growing urban populations\cite{xenochristou2020ensemble}.

In addition to operational benefits, consumption forecasting is essential for sustainability and environmental protection  \cite{brentan2017hybrid}. It helps identify trends related to climate change, population growth, and urbanisation, factors that affect water conservation policies. By predicting usage patterns, authorities can implement targeted conservation efforts, promote water-efficient technologies, and ensure the sustainable use of resources. This proactive approach helps mitigate the impacts of droughts and other water crises, ultimately enhancing urban resilience \cite{Rasifaghihi2020}.

\subsection{Related Work}\label{sec:related}
Traditionally, water demand forecasting methods have relied on statistical techniques such as Autoregressive Integrated Moving Average (ARIMA) and Seasonal ARIMA (SARIMA) models, which use historical consumption data to predict future demand \cite{herrera2010predictive}. However, these methods often struggle with the nonlinear and non-stationary characteristics of demand \cite{alvisi2007short}.

Machine learning models (e.g., neural networks, Support Vector Machines, and Random Forests) have been explored for their ability to capture complex, nonlinear relationships between water demand and its influencing factors, such as weather conditions, population growth, type of consumers, economic activities, water pricing, and water conservation policies \cite{adamowski2012comparison,ghalehkhondabi2017water}. The work in \cite{eliades2012leakage} proposed online learning to adapt to the non-stationary nature of DMA flow estimation, used for leakage detection. Deep learning techniques, such as Long Short-Term Memory (LSTM) networks have proven effective in handling sequential data and capturing both short- and long-term dependencies \cite{wang2024short}.

Recent works have highlighted the benefits of integrating real-time data to capture sudden changes in consumption patterns and enable more dynamic, responsive forecasts, particularly by considering the influence of external factors such as weather conditions and socio-economic changes \cite{zanfei2022short,wang2024short}.

\subsection{Contributions}

DMA flow forecasting is typically performed using data from the specific DMA. However, relying solely on historical data from a single DMA may not provide sufficient insight, as only localised information is considered. In addition, there are instances where a DMA is not monitored due to cost constraints, or where monitoring is interrupted by sensor malfunctions or faults. In such cases, forecasting based solely on historical data becomes challenging or even impossible. The contributions of this work are as follows:

\begin{enumerate}
    \item We propose a novel method in which given a DMA of interest, it first identifies correlated DMAs in terms of their consumption patterns, and second, as input to a deep learning model it uses its local and/or correlated consumption patterns to forecast the consumption of the DMA of interest.

    \item We conduct a real-world study concerning the city of Limassol, Cyprus; data corresponding to five DMAs were provided by the Water Board of Limassol.
\end{enumerate}

Our findings demonstrate that: i) The proposed deep learning model significantly outperforms a classical statistical model for DMA forecasting; ii) Forecasting a DMA of interest can be achieved by only considering the consumption pattern of its correlated DMAs; and iii) even if information about itself is available, it's sometimes beneficial for the deep learner to consider the correlated DMAs too.

The rest of the paper is structured as follows. Section~\ref{sec:formulation} formulates the problem and discusses the proposed method. Sections~\ref{sec:exp_setup} and \ref{sec:exp_results} provide the experimental setup and results respectively. Section~\ref{sec:conclusion} concludes the paper.

\section{Proposed Method}\label{sec:formulation}

\subsection{Problem formulation}
A WDN is defined as a network, with nodes representing junctions and edges representing pipes. A WDN is comprised of well-defined and isolated regions which are referred to as DMAs. Some DMAs may be monitored by the relevant authorities / network operators \cite{Vrachimis2018,Vrachimis2021}. For each monitored DMA, there is a sensor that records the total water consumption at every time step. We consider a data generating process that provides at each time step $t$ a sequence of instances $S = \{x^t\}_{t=1}^T \in \mathbb{R}^{n \times T}$ from an unknown probability distribution, where $n$ is the number of DMAs and $T \in \mathbb{Z}^+$ is the total number of time steps. The term $x^t = \{x^t_v\}_{v=1}^n \in \mathbb{R}^n$ is a $n$-dimensional vector which corresponds to the consumption of all DMAs at a given time $t$. The term $x_{v} = \{x_v^t\}_{t=1}^T \in \mathbb{R}^T$ is a $T$-dimensional vector, which corresponds to the consumption of a given DMA at node $v$, at all steps. The term $x^t_i \in \mathbb{R}$ denotes the current value of the DMA $i$ at time $t$.

\subsection{DMA consumption forecasting}

\textbf{Forecasting}. To capture the temporal aspects of the data, we introduce a memory component, specifically, a sliding window of size $W$ to facilitate the prediction task. At a node of interest $v$, a regression model is applied which observes a new example $x^t_v \in \mathbb{R}$ at time step $t$, as well as $W-1$ historical examples. Let $X_v^t = \{x_v^t, x_v^{t-1}, ..., x_v^{t-W+1}\}$, the regression model is defined by $h_{\textbf{local}}: \mathbb{R}^W \mapsto \mathbb{R}$ such that:
\begin{equation}\label{eq:single}
\hat{y}_v^{t+1} = h_{\text{local}}(X_v^t)
\end{equation}

\textbf{Training}. Following the effectiveness of deep learning in the field (see Sec.~\ref{sec:related}), we also propose to use a neural network for $h_{\text{local}}$.
We allow $W$ time steps to pass so that the sliding window becomes full, and the training set is provided by $X_v^{train} = \{X^t_v\}_{t=W}^{T_{train}}$. The cost function used is the Mean Squared Error (MSE) as follows:
\begin{equation}\label{eq:mse}
    J = \frac{1}{|X_v^{train}|} \sum_{t=W}^{T_{train}-1} (y_v^{t+1} - \hat{y}_v^{t+1})^2,
\end{equation}
\noindent where $y^{t+1}$ is the ground truth. Learning is performed using stochastic gradient descent (or any variant) where each neural network weight $w$ is updated according to the formula $w \leftarrow w - \alpha \frac{\partial J}{w}$ , where $\frac{\partial J}{w}$ is the partial derivative of $J$ with respect to $w$, and $\alpha$ is the learning rate. It is trained using the backpropagation algorithm.

\subsection{Consumption forecasting with corrrelated DMAs}

\textbf{Correlated DMAs}. To further improve the DMA prediction at node \( v \), we consider the current and historical consumption of other DMAs. To determine which DMAs will contribute to the prediction task, we calculate the correlation between the consumption patterns of different DMAs.

The correlation between two time series can be quantified using the Pearson correlation coefficient \cite{pearson1896vii}. For two DMAs, \( v_i \) and \( v_j \), with respective time series \( x_{v_i} \) and \( x_{v_j} \), the Pearson correlation coefficient $\rho_{v_i, v_j} \in [-1,1]$ is given by:

\begin{equation}\label{eq:p_corr}
\rho_{v_i, v_j} = \frac{\sum_{t=1}^{T_{train}} \left( x_{v_i}^t - \bar{x}_{v_i} \right) \left( x_{v_j}^t - \bar{x}_{v_j} \right)}{\sqrt{\sum_{t=1}^{T_{train}} \left( x_{v_i}^t - \bar{x}_{v_i} \right)^2} \sqrt{\sum_{t=1}^{T_{train}} \left( x_{v_j}^t - \bar{x}_{v_j} \right)^2}}
\end{equation}

Here, \( T_{train} \) represents the total number of observations in the training set, \( x_{v_i}^t \) and \( x_{v_j}^t \) represent the consumption values at time \( t \) for DMAs \( v_i \) and \( v_j \), respectively, and \( \bar{x}_{v_i} \) and \( \bar{x}_{v_j} \) are the mean values of the respective time series. A value close to \(1\) indicates a strong positive correlation, while a value close to \(-1\) indicates a strong negative correlation. A value around \(0\) suggests no linear correlation between the two time series.

The correlation set $C_v \subset V$ of a given DMA at node \( v \in V\) is defined as the set of nodes which are highly correlated with $v$. To choose the set of highly correlated DMAs based on a pre-defined threshold \( \theta \in [0,1] \), we define:
\begin{equation}\label{eq:corr_set}
C_v = \{v_j \mid \rho(v, v_j) \geq \theta \quad \forall v_j \in V, v \neq v_j\}
\end{equation}

In this context, \( \theta \) serves as a threshold to filter out DMAs that do not exhibit strong enough correlations with the DMA at node \( v \). Only DMAs with a correlation coefficient equal to or greater than \( \theta \) are included in the set $C_v$.

If the correlation coefficient \( \rho_{v, v_j} \) is significantly different from zero, it indicates that the consumption data from \( v_j \) can provide useful information for predicting the consumption at \( v \). This approach allows us to leverage the historical consumption data from correlated DMAs to improve the accuracy of predictions at a given node.

\begin{table*}[t]
\centering
\caption{DMA information}
\label{tab:DMA_info}
\resizebox{\textwidth}{!}{%
\begin{tabular}{|c|c|c|c|c|c|c|}
\hline
\textbf{DMA} & \textbf{Total area} ($km^2$) & \textbf{Piping length} (km) & \textbf{Num. of households} & \textbf{Mean daily water demand ($m^3$)}                  & \textbf{Neighbouring DMAs} & \textbf{Pressure zone} \\ \hline
\textbf{1}   & 0.76                          & 14.8                         & 5216                      & 23755                                                              & 2                           & 1            \\ \hline
\textbf{2}   & 0.7                           & 13.1                         & 790                       & 9803                                                   & 1                           & 1            \\ \hline
\textbf{3}   & 1.11                          & 24.5                         & 1600                      & 9028                                        & None                         & 2            \\ \hline
\textbf{4}   & 0.92                          & 21.5                         & 2924                      & 14897                                                                & 5                           & 2            \\ \hline
\textbf{5}   & 1.51                          & 31.5                         & 4663                      & 21898                                                               & 4                           & 2            \\ \hline
\end{tabular}%
}
\end{table*}

\textbf{Forecasting with correlated DMAs}. 
For a target DMA at node $v$, we now consider its consumption pattern based on a sliding window of size W, as well as the consumption patterns of correlated DMAs. Let $m = |C_v|$, then the regression model is defined by $h_{\textbf{local+correlated}}: \mathbb{R}^{(m + 1) \times W} \rightarrow \mathbb{R}$ such that:
\begin{equation}\label{eq:all}
\hat{y}^{t+1}_v = h_{\text{local+correlated}}(X_v^t, X_{v_1}^t, ..., X_{v_m}^t)
\end{equation}
\noindent where $v_1, ..., v_m \in C_v$ are the correlated DMAs. Without any loss of generality, we consider the same sliding window W.

Interestingly, provided that the correlated DMAs have already been determined (e.g., using historical / training data), we propose to forecast the consumption of a DMA of interest based only on the consumption patterns of correlated DMAs.
In this case, the regression model is defined by $h_{\textbf{correlated}}: \mathbb{R}^{m \times W} \rightarrow \mathbb{R}$ such that:
\begin{equation}\label{eq:correlated}
\hat{y}^{t+1}_v = h_{\text{correlated}}(X_{v_1}^t, ..., X_{v_m}^t)
\end{equation}

This approach ensures continuity in water consumption predictions, even in the absence of direct sensor data for the target DMA, by leveraging the DMA correlations within the water distribution network. Lastly, as before we propose to use a neural network for $h_{\text{local+correlated}}$ and $h_{\text{correlated}}$, and the training is performed in an analogous way as described earlier.

\begin{figure}[t]
    \centering
    \includegraphics[width=0.2\textwidth]{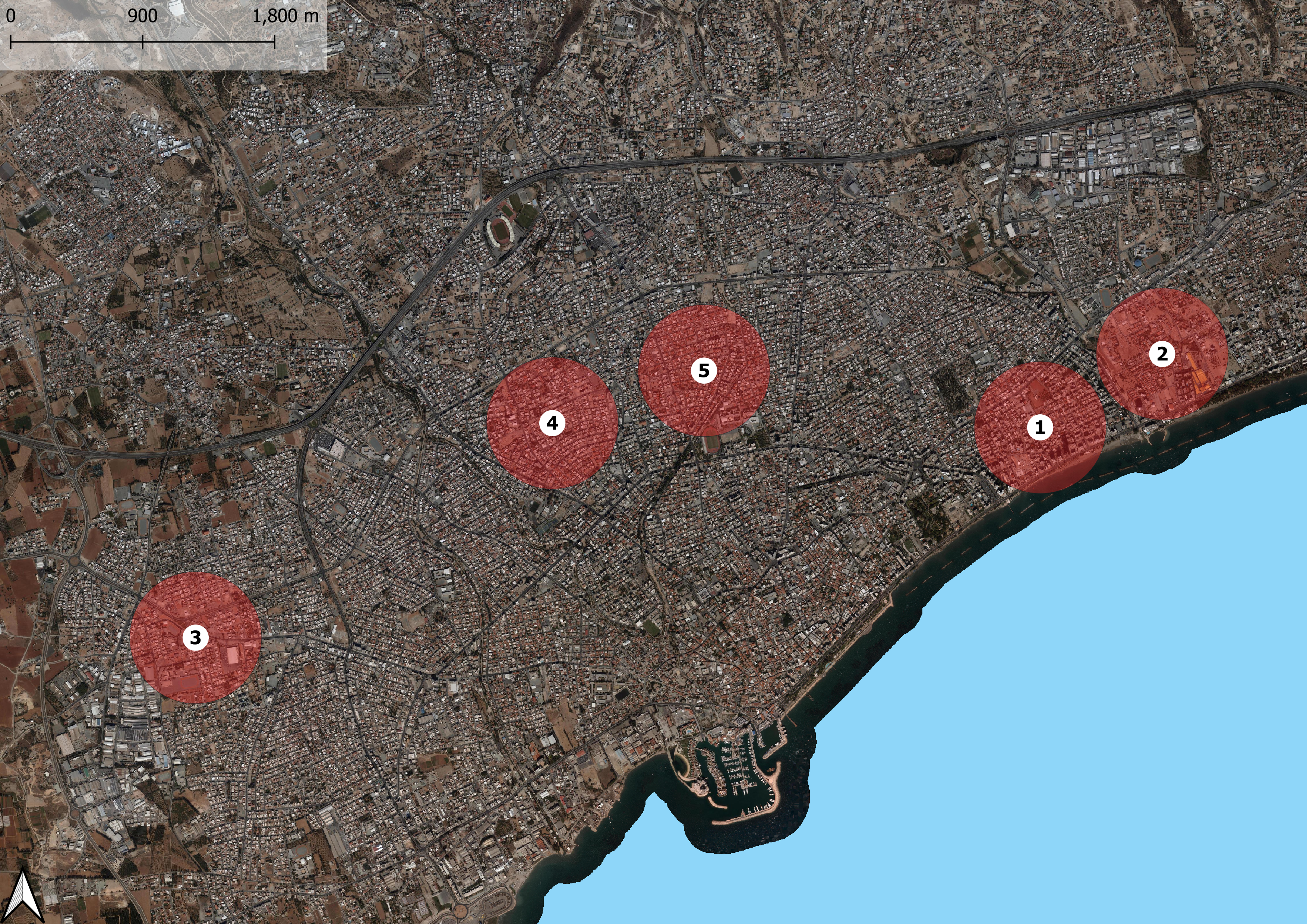} 
    \caption{Geographical location of the 5 DMAs}
    \label{fig:example_image}
\end{figure}

\section{Experimental setup}\label{sec:exp_setup}

\subsection{Dataset}
The Water Board of Limassol water supply network serves about 110,000 households (i.e., the number of installed water meters), corresponding to roughly 180,000 residents, and has an estimated annual average water demand of 17 million m³.
The DMAs selected for this study, for which time series data from flow sensors at 5-minute intervals were provided, are 1, 2, 3, 4, and 5 as shown in Fig.~\ref{fig:example_image}. Details regarding each DMA are provided in Table~\ref{tab:DMA_info}. Public data associated with population density and building use, are shown in Table~\ref{tab:DMA_build}. Low residential buildings are structures with up to three stories, while high residential buildings with greater population density, include hotels and other multi-story structures.

\subsection{Prediction models}
\textbf{SARIMA} \cite{box2015time}. It is a classical statistical forecasting method. While it is often considered a baseline method, it generally performs well across many domains, particularly in the case of limited data availability. 


\begin{table}[]
\centering
\caption{DMA building types}
\label{tab:DMA_build}
\begin{tabular}{|l|c|c|c|c|c|}
\hline
\textbf{Building type} & \textbf{1} & \textbf{2} & \textbf{3} & \textbf{4} & \textbf{5} \\ \hline
Low residential     & 560  & 198  & 1180 & 1043 & 2348  \\ \hline
High residential    & 329  & 57   & 177  & 368  & 536   \\ \hline
Commercial          & 3    & 5    & 9    & -     & 1     \\ \hline
Social              & -     & 1    & 4    & -      & 5     \\ \hline
Industrial          & -     & 31   & 8    &  -    &   -    \\ \hline
\end{tabular}
\end{table}

\textbf{CNN-LSTM}. The proposed deep learning model suitable for capturing temporal correlations in time series data. The network architecture leverages a combination of Convolutional Neural Networks (CNNs) \cite{lecun1998gradient} and recurrent neural networks, specifically GRU \cite{cho2014learning} and LSTM \cite{hochreiter1997long}, layers. To avoid overfitting we also use Dropout \cite{srivastava2014dropout} and L2 regularisation.
In all experiments, the window size is set to $W=15$.

\subsection{Evaluation methodology}
We use the Mean Squared Error (MSE), Mean Absolute Error (MAE), and Root Mean Squared Error (RMSE) metrics. We have considered the first 1.5 months as the train set, the following 0.5 months as the validation set, and the last 1 month as the test set. The learning models' results were repeated five times, and we present in Tables the mean and standard deviation, and in Figures the standard error around the mean.




\begin{table}[t]
\centering
\caption{DMA correlation matrix ($\geq 0.95$ in bold)}
\label{tab:DMA_corr}
\begin{tabular}{|c|c|c|c|c|c|}
\hline
             & \textbf{1}   & \textbf{2} & \textbf{3}   & \textbf{4}   & \textbf{5}   \\ \hline
\textbf{1} & \textbf{1}     & 0.621        & 0.923          & 0.916          & \textbf{0.951} \\ \hline
\textbf{2} & 0.621          & \textbf{1}   & 0.559          & 0.693          & 0.680          \\ \hline
\textbf{3} & 0.923          & 0.559        & \textbf{1}     & 0.906          & \textbf{0.950} \\ \hline
\textbf{4} & 0.916          & 0.693        & 0.906          & \textbf{1}     & \textbf{0.953} \\ \hline
\textbf{5} & \textbf{0.951} & 0.680        & \textbf{0.950} & \textbf{0.953} & \textbf{1}     \\ \hline
\end{tabular}
\end{table}

\section{Experimental results}\label{sec:exp_results}

\subsection{Related DMAs}
We calculate the correlation matrix using the train set (1.5 months); Table~\ref{tab:DMA_corr} shows the correlations. A threshold of 0.95 was used to determine significant correlations between the DMAs. The following are found to be strongly correlated: DMAs 4 and 5, DMAs 1 and 5, and DMAs 3 and 5.


Identifying these related DMAs can help in understanding the underlying factors affecting water usage and in designing more effective water management strategies. DMAs might exhibit correlated water consumption profiles due to several shared factors, such as, geographic proximity, shared infrastructure, similar consumer types, socioeconomic factors, population density, building types, and seasonal tourism.

Using public data associated with population density and building use for the DMAs studied, as shown in Table~\ref{tab:DMA_build}, we can make the following observations that likely contribute to the correlated consumption profiles presented in Table~\ref{tab:DMA_corr}:
\begin{itemize}
    \item \textbf{DMAs 4 and 5}: They have a large number of \textit{low residential buildings}, likely contributing to similar consumption patterns. Also, both are in the same pressure zone.
    \item \textbf{DMAs 1 and 5}: Both DMAs are characterised by a strong presence of \textit{low residential buildings}. Despite differences in total building counts, the predominance of suburban-style housing in these DMAs suggests correlated water usage trends focused on residential consumption.
    \item \textbf{DMAs 3 and 5}: DMA 3 has both \textit{high} and \textit{low residential buildings}, while low residential buildings dominate DMA 5. This suggests a potential correlation in consumption. Also, both DMAs are in the same pressure zone.
\end{itemize}


\begin{table}[t!]
\centering
\caption{Forecasting for DMA 5 by considering its own time series (``self'') and of correlated DMAs}
\label{tab:DMA5_self}
\begin{tabular}{|c|c|c|c|c|c|}
\hline
              & self (5) & self, 1, 3 & self, 1, 4 & self, 3, 4 & \textbf{self, 1, 3, 4} \\ \hline
\textbf{MSE}  & 17.5619      & 15.7642        & 18.0943        & 15.9575        & \textbf{15.1426}            \\ \hline
\textbf{MAE}  & 2.9813       & 2.93         & 3.2757         & 3.0193         & \textbf{2.9011}             \\ \hline
\textbf{RMSE} & 4.1907       & 3.9704         & 4.2537         & 3.9947         & \textbf{3.8914}             \\ \hline
\end{tabular}
\end{table}


\begin{table}[t]
\centering
\caption{Performance for DMA 5 by considering only the time series of correlated DMAs}
\label{tab:DMA5_corr}
\begin{tabular}{|c|c|c|c|c|c|c|c|}
\hline
              & 1, 3  & 1, 4  & 3, 4  & \textbf{1, 3, 4} \\ \hline
\textbf{MSE}  & 29.4398   & 31.8711   & 973.045   & \textbf{16.4295}       \\ \hline
\textbf{MAE}  & 3.9952    & 4.1131    & 24.9813   & \textbf{2.9946}        \\ \hline
\textbf{RMSE} & 5.4258    & 5.6455    & 31.1937   & \textbf{4.0533}        \\ \hline
\end{tabular}
\end{table}


\begin{table}[t]
\centering
\caption{Comparison for DMA 5}
\label{tab:comparison5}
\begin{tabular}{|c|c|c|c|}
\hline
              & SARIMA & self (5) & \textbf{self + corr. (1, 3, 4)} \\ \hline
\textbf{MSE}  & 1733.056 (1.7562)        & 17.5619 (1.456)                            & \textbf{15.1426  (0.0856)}            \\ \hline
\textbf{MAE}  & 37.1182 (0.4472)         & 2.9813 (0.389)                              & \textbf{2.9011 (0.0130)}              \\ \hline
\textbf{RMSE} & 41.63 (0.8381)           & 4.1907 (0.712)                             & \textbf{3.8914 (0.0145)}              \\ \hline
\end{tabular}
\end{table}

\subsection{Performance with self and correlated DMAs}\label{sec:exp_results_self_corr}
We now examine the forecasting performance of each DMA by using its own time series data (``self'') and the time series data from correlated DMAs. The correlations were established based on the correlation matrix provided earlier. Table~\ref{tab:DMA5_self} presents the MSE, MAE, and RMSE for DMA 5. The best performance is achieved when DMA 5's time series is combined with those of DMAs 1, 3, and 4. This suggests that incorporating the information from these correlated DMAs can enhance the accuracy of predictions for DMA 5.

\subsection{Performance with correlated DMAs only}\label{sec:exp_results_corr}

In this section, we examine the forecasting performance for each DMA when considering only the time series data from its correlated DMAs, excluding its own time series data. Table~\ref{tab:DMA5_corr} presents the results for DMA 5. The best performance is observed when combining all three correlated DMAs (1, 3, and 4). Interestingly, using DMA 3 alone results in significantly higher error metrics, demonstrating that some DMAs may improve the prediction when used in combination with others.

\subsection{Comparison}

In this section, we compare the performance of different forecasting approaches for DMAs. The methods compared include SARIMA, using the target DMA's own time series data (``self'') from Sec.~\ref{sec:exp_results_self_corr}, using only the correlated DMAs' time series (Sec.~\ref{sec:exp_results_corr}), and combining both the target DMA and its correlated DMAs' time series (Sec.~\ref{sec:exp_results_self_corr}).

Table~\ref{tab:comparison5} presents the results for DMA 5. The lowest error metrics (MSE, MAE, and RMSE) are observed when using both the target DMA's own time series along with the correlated DMAs (1, 3, and 4). This indicates that the combination of self and correlated DMAs provides the most accurate forecasting for DMA 5. In contrast, the SARIMA model performs significantly worse, with much higher error metrics, demonstrating its limitations in this context.

Figs. 2a - 2b show the forecasted water consumption for DMA 5 at different time intervals, showcasing 288 consecutive points with a time gap of 5 minutes between consecutive points (i.e., 24 hours). This segmentation allows us to examine the forecast behavior over short time windows.

\begin{figure}[!t]
  \centering
  

    \begin{subfigure}{.25\textwidth}
        \centering
        \includegraphics[width=0.9\columnwidth]{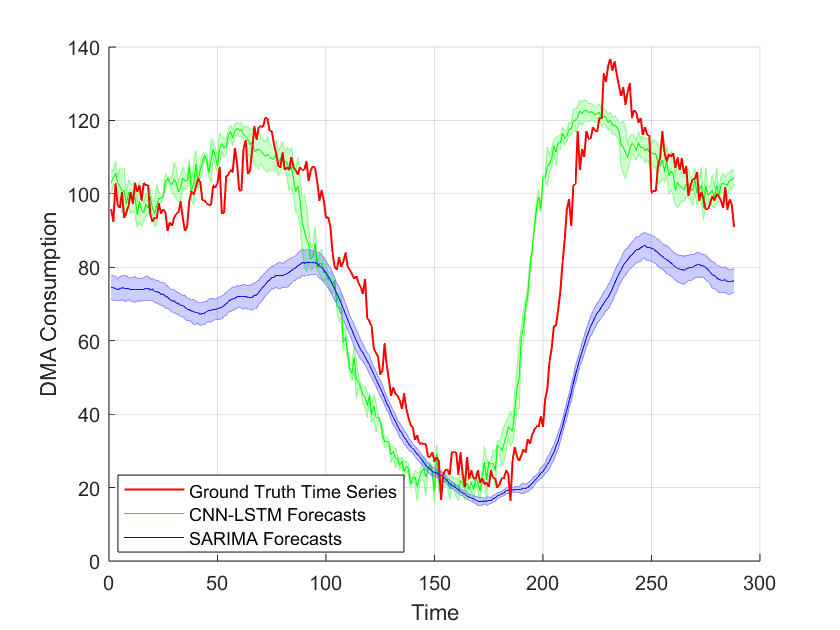} 
        \label{fig:exp_plot_230.3}
    \end{subfigure}%
    \begin{subfigure}{.25\textwidth}
        \centering
        \includegraphics[width=0.9\columnwidth]{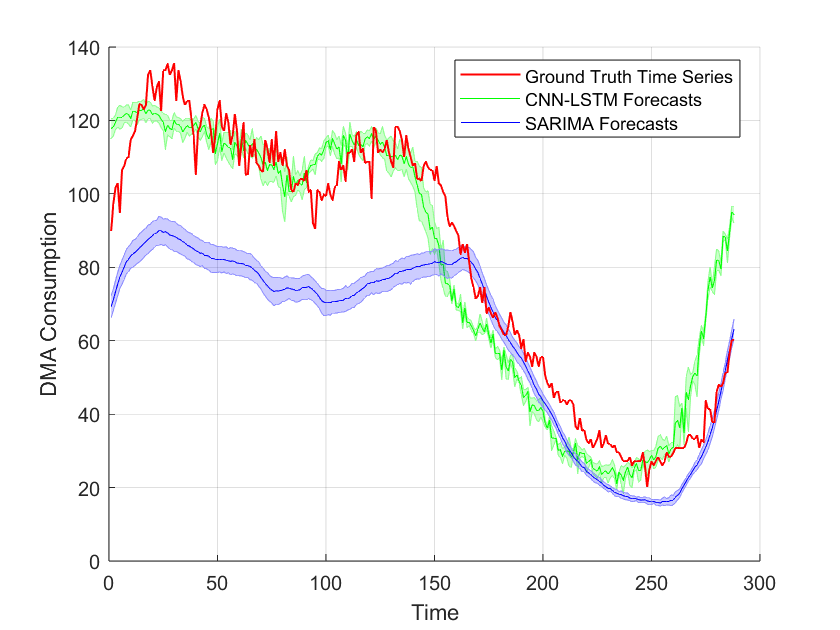} 
        \label{fig:exp_plot_230.6}
    \end{subfigure}%
    
    \caption{DMA 5 forecasting with SARIMA and CNN-LSTM (self + correlated) at different time intervals}
    \label{fig:exp}
\end{figure}

Overall, SARIMA exhibits smoother transitions in its forecasted values, reflecting a less reactive approach to short-term fluctuations in water demand. In contrast, the CNN-LSTM model captures the sharper variations (``ups'' and ``downs'') in water consumption more effectively, providing a forecast that aligns more closely with the observed real-world data.

\section{Conclusion}\label{sec:conclusion}
This work is concerned with DMA water consumption forecasting. It introduces a novel method which first identifies correlated consumption patterns from neighboring DMAs, then provides these to an advanced deep learning model. Considering any DMA of interest, the model offers a more accurate and reliable forecasting solution, even when the DMA faces sensor malfunctions or even when the specific DMA is not directly monitored due to computational costs.

\newpage

\bibliographystyle{IEEEtran}
\bibliography{camera_ready}

\begin{thebibliography}{10}
\providecommand{\url}[1]{#1}
\csname url@samestyle\endcsname
\providecommand{\newblock}{\relax}
\providecommand{\bibinfo}[2]{#2}
\providecommand{\BIBentrySTDinterwordspacing}{\spaceskip=0pt\relax}
\providecommand{\BIBentryALTinterwordstretchfactor}{4}
\providecommand{\BIBentryALTinterwordspacing}{\spaceskip=\fontdimen2\font plus
\BIBentryALTinterwordstretchfactor\fontdimen3\font minus
  \fontdimen4\font\relax}
\providecommand{\BIBforeignlanguage}[2]{{%
\expandafter\ifx\csname l@#1\endcsname\relax
\typeout{** WARNING: IEEEtran.bst: No hyphenation pattern has been}%
\typeout{** loaded for the language `#1'. Using the pattern for}%
\typeout{** the default language instead.}%
\else
\language=\csname l@#1\endcsname
\fi
#2}}
\providecommand{\BIBdecl}{\relax}
\BIBdecl

\bibitem{Fuertes2020}
M.~H.~C. P.~Conejos~Fuertes, F. Martínez~Alzamora and J.~A. Campos, ``Building
  and exploiting a digital twin for the management of drinking water
  distribution networks,'' \emph{Urban Water Journal}, vol.~17, no.~8, pp.
  704--713, 2020.

\bibitem{xenochristou2020ensemble}
M.~Xenochristou and Z.~Kapelan, ``An ensemble stacked model with bias
  correction for improved water demand forecasting,'' \emph{Urban Water
  Journal}, vol.~17, no.~3, pp. 212--223, 2020.

\bibitem{brentan2017hybrid}
B.~M. Brentan, G.~Meirelles, D.~Manzi, M.~Herrera, J.~Izquierdo, and
  E.~Luvizotto, ``Hybrid regression model for near real-time urban water demand
  forecasting,'' \emph{Journal of Computational and Applied Mathematics}, vol.
  309, pp. 532--541, 2017.

\bibitem{Rasifaghihi2020}
N.~Rasifaghihi, S.~Li, and F.~Haghighat, ``Forecast of urban water consumption
  under the impact of climate change,'' \emph{Sustainable Cities and Society},
  vol.~52, p. 101848, 2020.

\bibitem{herrera2010predictive}
M.~Herrera, L.~Torgo, J.~Izquierdo, and R.~P{\'e}rez-Garc{\'\i}a, ``Predictive
  models for forecasting hourly urban water demand,'' \emph{Journal of
  Hydrology}, vol. 387, no. 1-2, pp. 141--150, 2010.

\bibitem{alvisi2007short}
S.~Alvisi, M.~Franchini, and A.~Marinelli, ``A short-term, pattern-based model
  for water-demand forecasting,'' \emph{Journal of Hydroinformatics}, vol.~9,
  no.~1, pp. 39--50, 2007.

\bibitem{adamowski2012comparison}
J.~Adamowski, H.~Fung~Chan, S.~O. Prasher, B.~Ozga-Zielinski, and
  A.~Sliusarieva, ``Comparison of multiple linear and nonlinear regression,
  autoregressive integrated moving average, artificial neural network, and
  wavelet artificial neural network methods for urban water demand forecasting
  in montreal, canada,'' \emph{Water Resources Research}, vol.~48, no.~1, 2012.

\bibitem{ghalehkhondabi2017water}
I.~Ghalehkhondabi, E.~Ardjmand, W.~A. Young, and G.~R. Weckman, ``Water demand
  forecasting: review of soft computing methods,'' \emph{Environmental
  Monitoring and Assessment}, vol. 189, pp. 1--13, 2017.

\bibitem{eliades2012leakage}
D.~Eliades and M.~M. Polycarpou, ``Leakage fault detection in district metered
  areas of water distribution systems,'' \emph{Journal of Hydroinformatics},
  vol.~14, no.~4, pp. 992--1005, 2012.

\bibitem{wang2024short}
D.~Wang, Y.~Li, B.~Hou, and S.~Wu, ``Short-term water demand forecasting based
  on lstm using multi-input data,'' \emph{Engineering Proceedings}, vol.~69,
  no.~1, p. 103, 2024.

\bibitem{zanfei2022short}
A.~Zanfei, B.~M. Brentan, A.~Menapace, and M.~Righetti, ``A short-term water
  demand forecasting model using multivariate long short-term memory with
  meteorological data,'' \emph{Journal of Hydroinformatics}, vol.~24, no.~5,
  pp. 1053--1065, 2022.

\bibitem{Vrachimis2018}
S.~G. Vrachimis, D.~G. Eliades, and M.~M. Polycarpou, ``{Real-time hydraulic
  interval state estimation for water transport networks: a case study},''
  \emph{Drinking Water Engineering and Science}, vol.~11, no.~1, pp. 19--24,
  Mar 2018.

\bibitem{Vrachimis2021}
S.~G. Vrachimis, S.~Timotheou, D.~G. Eliades, and M.~M. Polycarpou, ``{Leakage
  detection and localization in water distribution systems: A model
  invalidation approach},'' \emph{Control Engineering Practice}, vol. 110, p.
  104755, May 2021.

\bibitem{pearson1896vii}
K.~Pearson, ``Vii. mathematical contributions to the theory of
  evolution.—iii. regression, heredity, and panmixia,'' \emph{Philosophical
  Transactions of the Royal Society of London. Series A, containing papers of a
  mathematical or physical character}, no. 187, pp. 253--318, 1896.

\bibitem{box2015time}
G.~E. Box, G.~M. Jenkins, G.~C. Reinsel, and G.~M. Ljung, \emph{Time series
  analysis: forecasting and control}.\hskip 1em plus 0.5em minus 0.4em\relax
  John Wiley \& Sons, 2015.

\bibitem{lecun1998gradient}
Y.~LeCun, L.~Bottou, Y.~Bengio, and P.~Haffner, ``Gradient-based learning
  applied to document recognition,'' \emph{Proceedings of the IEEE}, vol.~86,
  no.~11, pp. 2278--2324, 1998.

\bibitem{cho2014learning}
K.~Cho, B.~van Merrienboer, C.~Gulcehre, D.~Bahdanau, F.~Bougares, H.~Schwenk,
  and Y.~Bengio, ``Learning phrase representations using rnn encoder-decoder
  for statistical machine translation,'' in \emph{Conference on Empirical
  Methods in Natural Language Processing (EMNLP)}, 2014.

\bibitem{hochreiter1997long}
S.~Hochreiter, ``Long short-term memory,'' \emph{Neural Computation MIT-Press},
  1997.

\bibitem{srivastava2014dropout}
N.~Srivastava, G.~Hinton, A.~Krizhevsky, I.~Sutskever, and R.~Salakhutdinov,
  ``Dropout: a simple way to prevent neural networks from overfitting,''
  \emph{The Journal of Machine Learning Research}, vol.~15, no.~1, pp.
  1929--1958, 2014.

\end{thebibliography}

\end{document}